\documentclass[letterpaper, 10 pt, conference]{ieeeconf}  

\IEEEoverridecommandlockouts                              
\usepackage{hyperref}
\usepackage{cite}
\usepackage{graphicx} 
\usepackage{epstopdf}
\usepackage{amsfonts}
\usepackage{amsmath}
\usepackage{multicol}  
\usepackage{multirow}

\title{\LARGE \bf
Retrieval-based Localization Based on Domain-invariant Feature Learning under Changing Environments
}

\author{Hanjiang Hu, Hesheng Wang$^{*}$, Zhe Liu, Chenguang Yang, Weidong Chen, and Le Xie
\thanks{This work was supported in part by the Natural Science Foundation of China under Grant U1613218, 61722309 and U1813206, in part by State Key Laboratory of Robotics and System (HIT). Corresponding Author: Hesheng Wang.   }
\thanks{H. Hu, H. Wang and W. Chen are with Autonomous Robot Lab, Department of Automation, Key Laboratory of System Control and Information Processing of Ministry of Education, Shanghai Jiao Tong University, Shanghai 200240, China. Hesheng Wang is also with the State Key Laboratory of Robotics and System (HIT). Z. Liu is with the T Stone Robotics Institute, Chinese University of Hong Kong, Hong Kong, China. C. Yang is with Bristol Robotics Laboratory, University of the West of England, Bristol, BS16 1QY, UK. L. Xie is with the School of Material Science and Engineering, Shanghai Jiao Tong University, Shanghai 200240, China.}
}

\begin{document}

\maketitle
\thispagestyle{empty}
\pagestyle{empty}

\begin{abstract}

Visual localization is a crucial problem in mobile robotics and autonomous driving. One solution is to retrieve images with known pose from a database for the localization of query images. However, in environments with drastically varying conditions (e.g. illumination changes, seasons, occlusion, dynamic objects), retrieval-based localization is severely hampered and becomes a challenging problem. In this paper, a novel domain-invariant feature learning method (DIFL) is proposed based on ComboGAN, a multi-domain image translation network architecture. By introducing a feature consistency loss (FCL) between the encoded features of the original image and translated image in another domain, we are able to train the encoders to generate domain-invariant features in a self-supervised manner. To retrieve a target image from the database, the query image is first encoded using the encoder belonging to the query domain to obtain a domain-invariant feature vector. We then preform retrieval by selecting the database image with the most similar domain-invariant feature vector. We validate the proposed approach on the CMU-Seasons dataset, where we outperform state-of-the-art learning-based descriptors in retrieval-based localization for high and medium precision scenarios.

\end{abstract}

\section{INTRODUCTION}
\label{sec1}
Visual localization, an essential problem in computer vision, is widely used in many applications such as autonomous mobile robotics and self-driving vehicles. Given a database of images taken under the same conditions (e.g. illumination, season, time-of-day, etc.) and their corresponding poses, it is intuitive and effective to localize a query image taken under different conditions using image retrieval, i.e. place recognition. This retrieval-based technique is widely used in SLAM and loop closure detection\cite{gao2018ldso}.

Retrieval-based localization faces several challenges when applied in robotics and self-driving, mostly owing to the changing environmental conditions. The visual variability caused by different seasons, varying illumination, shifting perspectives, and dynamic objects significantly influences the quality of visual place recognition.
 
\begin{figure}[thpb]
	\centering
	\includegraphics[scale=0.32]{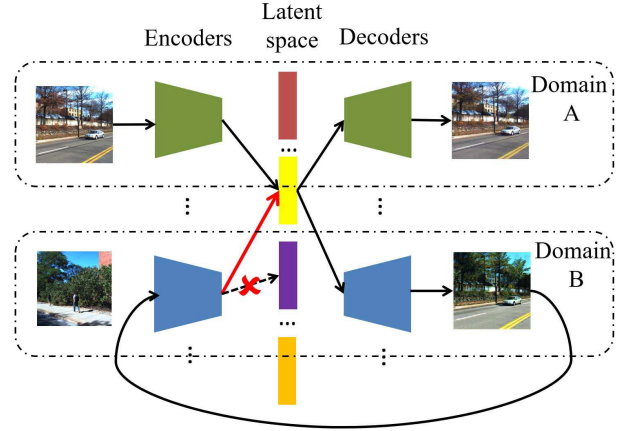}
	\caption
	{
	Proposed generators for multi-domain images are divided into encoders and decoders for each domain, with the latent space composed of encoded features shared among domains. We desire each encoded feature to be solely specified by the corresponding place and not related to any domain, creating so-called domain-invariant features. To achieve this, while training image translation from domain $ A $ to domain $ B $, we propose a method to compel the domain $ B $ encoder to encode the translated domain $ A $ image the same way as the real domain $ A $ image is encoded (shown by the red arrow), instead of any other encoded feature in latent space (shown by the red cross). This is implemented through a loss called feature consistency loss (FCL) with details in Section \hyperref[sec3sub2]{\ref{sec3sub2}} .
	}
	\label{fig1}
\end{figure}
Under a static scene, place recognition has been addressed successfully through using local features (SIFT, SUFT, ORB, etc.) and global features of the image. These man-made feature descriptors show satisfactory invariance under changing perspectives and moderate occlusion. However, these approaches for place recognition work poorly in dynamic environments and changing conditions. With deep neural networks making great progress in computer vision, learning-based features have shown remarkable advantages in place recognition in these dynamic environments, resulting in more robust outputs and semantic features from CNNs for example.

Unlike other tasks of recognition in computer vision (e.g. face recognition) it is difficult to use supervised learning for place recognition due to a difficulty in determining which sets of images are classified as belonging to one scene. In particular, in situations where a series of images are taken in quick succession, it is difficult to manually determine which sets of subsequent images should be grouped together into one scene. To avoid this problem, unsupervised approaches have been proposed recently which aim to learn condition-invariant features. Lowry $ et\ al. $ \cite{lowry2016supervised} proposed a simple approach based on using modified PCA to remove dimensions of variant conditions and showed impressive results. Adversia Porav $ et\ al. $ \cite{porav2018adversarial} and Anoosheh $ et\ al. $ \cite{anoosheh2019night} both overcame condition variance through image translation. Yin $ et\ al. $ \cite{yin2019multi} proposed to separate condition-invariant features from extracted features using a CNN. In this work we propose a completely learning-based approach based on ComboGAN \cite{anoosheh2018combogan} to directly extract domain-invariant features with the generator training method illustrated in Figure \hyperref[fig1]{\ref{fig1}}. We outperform the state-of-the-art learning-based approach NetVLAD \cite{arandjelovic2016netvlad}, especially when foliage is present. In summary, our work makes the following contributions:

\begin{itemize}
\item We introduce a retrieval-based localization approach using domain-invariant features based on ComboGAN and propose a novel feature consistency loss for place recognition instead of just image-to-image translation among different domains.
\item We validate the effectiveness of DIFL and FCL through experimental comparison on the urban part of the CMU-Seasons dataset. 
\item We show that our approach outperforms the state-of-the-art learning-based approach NetVLAD \cite{arandjelovic2016netvlad} in high and medium precision regimes on the complete CMU-Seasons dataset.
\end{itemize}

The rest of this paper is structured as follows. Section \hyperref[sec2]{\ref{sec2}} analyzes the related work in feature representation and place recognition. Section \hyperref[sec3]{\ref{sec3}} introduces the proposed method. Section \hyperref[sec4]{\ref{sec4}} presents the experimental results. Finally, in Section \hyperref[sec5]{\ref{sec5}} we draw our conclusions and present some suggestions for future work.

\section{RELATED WORK}
\label{sec2}

\subsection{Image Translation}
\label{sec2sub1}

In recent years, the generative adversarial network (GAN) has garnered significant attention due to its impressive results as a generative model. It is a common problem to translate images from domain $A$ to domain $B$ in computer vision tasks (e.g. style transfer, etc.) Isola $ et\ al. $ \cite{isola2017image} proposed the first GAN-based approach for image to image translation, where the generator generates images from the properties of exiting images instead of from samples of feature vector distributions like classic GAN frameworks \cite{goodfellow2014generative}. However, it is a supervised learning framework that requires manually labeled image pairs.

CycleGAN \cite{zhu2017unpaired}, introduced by Zhu $ et\ al. $, utilizes the GAN framework in an unsupervised manner, without any alignment of image pairs. CycleGAN is composed of two pairs of networks, $(G, DA)$ and $(F, DB)$. The generators $G$ and $F$ translate from domain $A$ to $B$ and $B$ to $A$ respectively, while the discriminators $DA$ and $DB$ are able to distinguish real images $a$ and $b$ from translated images $F(a)$ and $G(b)$ respectively. It consists of both an adversarial loss and a cycle consistency loss while training.

Many works based on CycleGAN are proposed, e.g. Liu $ et\ al. $ \cite{liu2017unsupervised} implemented the CycleGAN architecture together with a variational-autoencoder loss on the shared latent space to improve the image translation. But both \cite{zhu2017unpaired} and \cite{liu2017unsupervised} only work for two domains per training process, which is not suitable for outdoor place recognition tasks. StarGAN \cite{choi2018stargan} is another unsupervised image-translation approach, which uses one generator and discriminator for all domains instead of multiple generators and discriminators as in CycleGAN. It solved the difficulty of multi-domains translation but was limited to the facial recognition application, where all the domains were distributed around one specific category with slight variance.

ComboGAN, proposed in \cite{anoosheh2018combogan}, extended CycleGAN to multiple domains but retains the framework of multiple generators and discriminators. It presents promising results for image-translation. Huang $ et\ al. $ proposed MUNIT \cite{huang2018multimodal}, which implements multi-modal image translation without deterministic domains or modes, showing impressive disentanglement of content and style.

\subsection{Place Recognition and Localization}
\label{sec2sub2}

Place recognition deals with finding the most similar database image for a query image, which can be regarded as the image retrieval for localization task. In loop closure detection in SLAM, the early methods focus on local feature descriptors, e.g. FAB-MAP \cite{cummins2008fab}. These generally perform well on the famous real-time V-SLAM systems \cite{mur2017orb}, but they fail if the query and database images are taken under tremendously different environments due to the mismatch of descriptors between them. DBoW2 \cite{galvez2012bags})SeqSLAM \cite{milford2012seqslam} uses sequence information to avoid such failures but brings concerns about failure under a large variance of perspectives. 

VLAD \cite{jegou2010aggregating} is the most widely used hand-crafted descriptor in place recognition. A VLAD descriptor is a global feature representation of the whole image, created by aggregating the sum of residuals between cluster centers and their local descriptors on every dimension. Based on VLAD, DenseVLAD \cite{torii201524} was proposed by by Torii $ et\ al. $, which extracts SIFT descriptors at different scales to represent the image with multiple VLAD versions. NetVLAD \cite{arandjelovic2016netvlad} uses convolutional networks to learn global features according to a VLAD-process-like network architecture. NetVLAD gives impressive results by replacing the traditional VLAD process with a neural network module.

Approaches using a combination of image-to-image translation and VLAD-like descriptors, have also been proposed recently for retrieval-based localization. Porav $ et\ al. $ \cite{porav2018adversarial} uses CycleGAN to do appearance transfer with a new cycle loss based on the SURF detector. ToDayGAN \cite{anoosheh2019night} implements modified CycleGAN/ComboGAN to translate images from night to day and uses DenseVLAD to retrieve images from database. 

\subsection{Feature Learning for Place Recognition}
\label{sec2sub3}

Recently, learning-based methods have drawn significant attention for place recognition and localization. \cite{lowry2016supervised} uses PCA for latent space embeddings generated by a pretrained classification network, where PCA removes the variance of environments while retaining condition-invariant features.
\begin{figure}[thpb]
	\centering
	\includegraphics[scale=1]{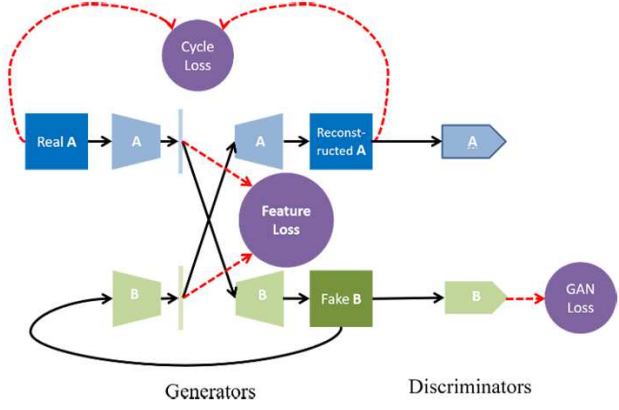}
	\caption
	{
	Architecture of translation from domains $ A $ to $ B $ where $ A $ and $ B $ are randomly selected. The pass from domains $ B $ to $ A $ is done in the same fashion. Overall, the generator training pass consists of three losses, represented as purple circles. Connections to the loss are marked as red dotted lines, while other connections in the pipeline are marked as black solid lines.
	}
	\label{fig2}
\end{figure}
 NetVLAD uses a CNN to extract features which are experimentally proven to be robust and independent to changing conditions. Yin $ et\ al. $ \cite{yin2019multi} recently proposed a multi-domain feature learning method, which first extracts VLAD features and then separate condition-invariant features from them using a GAN architecture.
 
 The multi-domain feature learning method \cite{yin2019multi} proposed by Yin $ et\ al. $ is the most similar method to ours. But compared our method of directly learning the features using a neural network, it uses a more complicated technique of first extracting VLAD descriptors and then separating them through neural networks.  Intuitively, ComboGAN's flexible combination of encoder-decoder pairs can effectively learn and extract domain-invariant features across multiple image domains. With this train of thought, we propose a novel, completely learning-based approach based on ComboGAN that shows great effectiveness in extracting domain-invariant features even under multiple environmental changes and is able to accomplish the retrieval-based localization task.

\section{PROPOSED METHOD}
\label{sec3}

In this section, we introduce domain-invariant feature learning (DIFL) based on the ComboGAN architecture and propose feature consistency loss (FCL) to keep the content of image embeddings identical across different domains, i.e. different styles of images. The procedure of image retrieval for localization is illustrated subsequently.

\subsection{Domain-invariant Feature Learning} 
\label{sec3sub1}

ComboGAN \cite{anoosheh2018combogan} successfully expanded the scale of CycleGAN \cite{zhu2017unpaired} from two to multiple domains through a decoupling of the generator networks into domain-specific sets of encoders and decoders. The first half of the generator is regarded as an encoder and the latter half is regarded as a decoder. These encoders and decoders can be manipulated as blocks due to the relationship of the corresponding domains. Then for a image taken at a specific place and pose, the extracted feature vector would be able to represent the specific place and pose, regardless of the environment that the image was taken under.

During each training iteration, two domains $ A, B \in \{1 \cdots N\} $ are selected randomly from the set of all domains, and two images $ a \sim p_{A}(a), b \sim p_{B}(b) $ are sampled from each domain respectively. For generators and discriminators trained in turn across domains, denote the encoder, decoder and discriminator for domain $ A $ as $ Ec_{A},Dc_{A} $ and $ D_{A} $ respectively. And let $ G_{A}(a) $ be short for $ Dc_{A}(Ec_{A}(a)) $ and $ G_{AB}(a) $ be short for $ Dc_{B}(Ec_{A}(a)) $. The basic ComboGAN architecture of \cite{anoosheh2018combogan} contains adversarial loss \cite{goodfellow2014generative} and cycle consistency loss \cite{zhu2017unpaired}, which can be formulated as Equation (\hyperref[gan_loss]{\ref{gan_loss}}) and Equation (\hyperref[cycle_loss]{\ref{cycle_loss}}) for translation from domain $A$ to $B$. This is illustrated as Figure \hyperref[fig2]{\ref{fig2}}.
\begin{align}
\label{gan_loss}
&\mathcal{L}_{GAN}(G_{AB},G_{B},A,B)=\mathbb{E}_{b \sim p_{B}(b)}[(D_{B}(b)-1)^{2}] \notag \\ &+\mathbb{E}_{a \sim p_{A}(a)}[D_{B}(G_{AB}(a))^{2}] \\  
\label{cycle_loss}    
&\mathcal{L}_{Cycle}(G_{AB},G_{BA})=\mathbb{E}_{a \sim p_{A}(a)}[\|G_{BA}(G_{AB}(a))-a\|_{1}] \notag\\
&+\mathbb{E}_{b \sim p_{B}(b)}[\|G_{AB}(G_{BA}(b))-b\|_{1}]       
\end{align}

In order to explain the domain-invariance of features in the latent space, we suppose that the ComboGAN networks are well trained with regards to minimizing the GAN loss and cycle consistency loss (i.e. for any domain and any image sample, image-to-image translation works without any concern). Now consider the case of translating image $ a $ from domain $ A $ into two different domains: from domain $ A $ to $ B $, and from domain $ A $ to domain $ C $, due to cycle consistency loss (\hyperref[cycle_loss]{\ref{cycle_loss}}), we have
\begin{eqnarray}
\label{equ3}
G_{BA}(G_{AB}(a))=G_{CA}(G_{AC}(a))=a  
\end{eqnarray}
Simpifying Equation (\hyperref[equ3]{\ref{equ3}}) by eliminating the deterministic probability distribution of $ Dc_{A} $, we have
\begin{eqnarray}
\label{equ4} 
&\forall A,B,C  \in \{1 \cdots  N\}, a \sim p_{A}(a), \notag \\
& Ec_{B}(Dc_{B}(Ec_{A}(a)))  =Ec_{C}(Dc_{C}(Ec_{A}(a)))  
\end{eqnarray}
Equation (\hyperref[equ4]{\ref{equ4}}) shows that for any image in any domain, the features encoded by the domain's corresponding encoder is independent of the domain itself, revealing the domain-invariance nature of the encoded feature vector.

\subsection{Feature Consistency Loss}
\label{sec3sub2}

Though the features generated from the original ComboGAN networks are domain-invariant and only depend on the place or content of the image, there is no explicit constraint on the content of translated fake image and original real image. Consequently, training the model to an ideal level with limited computational resources and time is somewhat challenging. To improve the training efficiency and to make the model more practical for the place recognition and localization task, we propose adding a feature consistency loss (\hyperref[feature_loss]{\ref{feature_loss}}), built on the encoded features of different domains and shown in Figure \hyperref[fig2]{\ref{fig2}}.
\begin{figure}[thpb]
	\centering
	\includegraphics[scale=1.0]{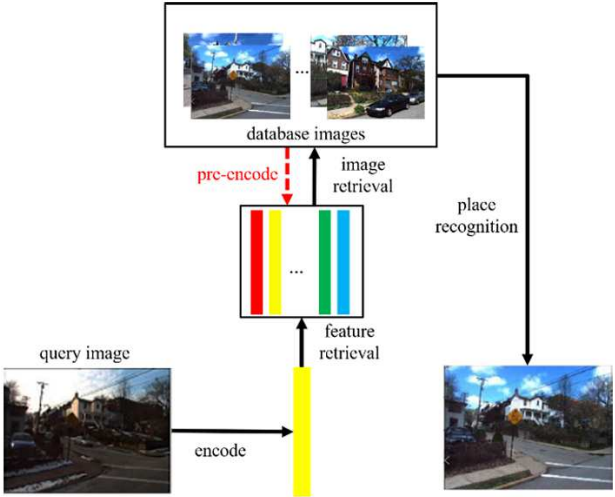}
	\caption
	{
		The place recognition process for retrieval-based localization. The query image is encoded and used to retrieve the most similar feature vector in the pre-encoded feature database, obtaining the corresponding reference image as a result. The database images are pre-encoded (represented by the red dotted line) into domain-invariant features in the middle, with every domain-invariant feature corresponding to a specific place, as denoted by the different colors. The place recognition proceduce follows the black solid lines.
	}
	\label{fig3}
\end{figure}
\begin{eqnarray}  
\label{feature_loss}   
\mathcal{L}_{Feature}(Ec_{A},Ec_{B},G_{AB},G_{BA})=\notag\\ \mathbb{E}_{a \sim p_{A}(a)}[\|Ec_{B}(G_{AB}(a))-Ec_{A}(a)\|_{2}] \\ 
+\mathbb{E}_{b \sim p_{B}(b)}[\|Ec_{A}(G_{BA}(b))-Ec_{B}(b)\|_{2}] \notag       
\end{eqnarray}
The feature consistency loss can be regarded as a kind of regularization term to make the model more robust and easier to train. Together with GAN loss and cycle consistency loss, the total loss is the sum of Equations (\hyperref[gan_loss]{\ref{gan_loss}}), (\hyperref[cycle_loss]{\ref{cycle_loss}}), (\hyperref[feature_loss]{\ref{feature_loss}}) weighted with hyperparameters $ \lambda_{1},\lambda_{2} $, as shown in Equation (\hyperref[total_loss]{\ref{total_loss}}).
\begin{align} 
\label{total_loss}    
&\mathcal{L}_{Total}(Ec_{A},Ec_{B},G_{AB},G_{BA},D_{A},D_{B})=\notag\\
&\mathcal{L}_{GAN}(G_{AB},G_{B},A,B)+\mathcal{L}_{GAN}(G_{BA},G_{A},B,A)+\\ 
&\lambda_{1}\mathcal{L}_{Cycle}(G_{AB},G_{BA})+\lambda_{2}\mathcal{L}_{Feature}(Ec_{A},Ec_{B},G_{AB},G_{BA})  \notag    
\end{align}

Suppose ComboGAN with FCL (\hyperref[feature_loss]{\ref{feature_loss}}) is well trained and the total loss (\hyperref[total_loss]{\ref{total_loss}}) is satisfied for any domain and image sample. Then, according to Equation (\hyperref[equ4]{\ref{equ4}}) we have
\begin{eqnarray} 
\label{equ7}    
&\forall A, I \in \{1 \cdots  N\}, a \sim p_{A}(a), \notag \\
& Ec_{I}(Dc_{I}(Ec_{A}(a)))=Ec_{A}(a)
\end{eqnarray}
Equation (\hyperref[equ7]{\ref{equ7}}) is further strengthened compared to Equation (\hyperref[equ4]{\ref{equ4}}), with the specification that given domain $ I $ and sample image $ a $ taken in domain $ A $, the result of $ Ec_{I}(Dc_{I}(Ec_{A}(a))) $ is not only independent of $ I $, but also only varies as a function of $ Ec_{A}(a) $. With some simplification, we see that the result is equal to $ Ec_{A}(a) $ itself, which is equivalent to the auto-encoder loss for the generator of each domain.

Figure \hyperref[fig2]{\ref{fig2}} shows that after randomly choosing two domains $ A, B $, the forward translation pass from $ A $ to $ B $, is essentially the same as as the pass from $ B $ to $ A $, but with the order of $ A $ and $ B $ exchanged. Note that total loss (\hyperref[total_loss]{\ref{total_loss}}) consists of both translations from $ A $ to $ B $ and from $ B $ to $ A $. 

\subsection{Image Retrieval Process}
\label{sec3sub3}

Our retrieval-based localization is based on domain-invariant feature learning. First, we train the networks described in Section \hyperref[sec3sub1]{\ref{sec3sub1}} and Section \hyperref[sec3sub2]{\ref{sec3sub2}} with images under changing environments. Then we use the trained model to pre-encode each database image into a one-dimentional vector to avoid redundant calculations when retrieving database images that correspond to the query image.

For every query image, we first use the corresponding trained encoder networks to extract features for the query image, then compare the feature with every feature vector in the database using a $ cosine\  distance $ metric (note that the metric used in Equation (\hyperref[feature_loss]{\ref{feature_loss}}) for training is $ L2 $, which will be discussed in Section \hyperref[sec4sub2]{\ref{sec4sub2}}). We choose the image with the most similar feature to be the retrieval result. Figure \hyperref[fig3]{\ref{fig3}} presents the place recognition process, where the query image is first encoded to be domain-invariant and then used to retrieve the feature and image with the largest similarity in the database.

\section{EXPERIMENTAL RESULTS}
\label{sec4}

We design a series of experiments to validate our domain-invariant feature learning retrieval approach and the effectiveness of feature consistency loss. And we compare our results with several localization baselines on CMU-Seasons dataset. We conduct experiments on two NVIDIA 1080Ti cards with 64G RAM on Ubuntu 16.04 system. Our source code and pre-trained models are available on \href{https://github.com/HanjiangHu/DIFL-FCL/}{\texttt{https://github.com/HanjiangHu/DIFL-FCL/}}. 

\subsection{Experimental Setup}
\label{sec4sub1}

The experiments are conducted on the CMU-Seasons dataset, which is presented in \cite{sattler2018benchmarking} and is based on the CMU Visual Localization \cite{Badino2011} dataset. It was recorded over the course of a year by having a vehicle with a left-side and a right-side camera drive on a 9 kilometers long route. The dataset is challenging due to the variance of environmental conditions as a result of changing seasons, illumination, weather, and especially foliage. The derived visual localization CMU-Seasons dataset is benchmarked in \cite{sattler2018benchmarking}, which gives a clear category and division of the original dataset, together with the groudtruth data for camera pose per reference database image. There are three areas and seventeen slices in the CMU-Seasons dataset: 31250 images in seven slices for the urban,13736 images in three slices for the suburban and 30349 images in seven slices for the park area. Additionally, there is one reference and eleven query conditions for each area. The database is under the condition of sunny with no foliage, while the query image can be chosen from sunny, cloudy, overcast, snow, etc. intersected with dense, mixed or no foliage.

Since our approach is unsupervised and the dataset lacks extra images for every condition, we use all the images as our training dataset and train the model separately for each area part. While testing, we follow the image retrieval process described in Section \hyperref[sec3sub3]{\ref{sec3sub3}} slice by slice to improve efficiency. 
\begin{figure}[thpb]
	\centering
	\includegraphics[scale=0.5]{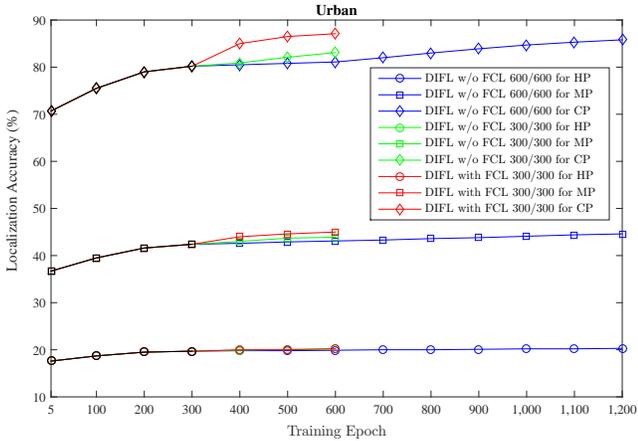}
	\caption
	{
		The localization results of three trials described in \hyperref[sec4sub2]{\ref{sec4sub2}} are illustrated during the whole training process. Blue lines are results 600-600 training without FCL while green lines are 300-300. Red lines shows results with FCL and mixed black lines are from the shared pre-trained model. Lines with circles, squares and diamonds represent results in regimes of high precision (HP), medium precision (MP) and coarse precision (CP) respectively.
	}
	\label{fig4}
\end{figure}
The images are scaled to $ 286 \times 286$ and cropped to $ 256 \times 256$ size randomly while training. And the dimension of encoded feature vector is flatted after the output of the fourth ResNetBlock with a shape of $ 256 \times 72 \times 96 $. Experiment with dimensionality reduction of features through PCA is discussed in Section \hyperref[sec4sub2]{\ref{sec4sub2}}.We evaluate the retrieval-based localization following the protocol introduced in \cite{sattler2018benchmarking}, which is the percentage of query images correctly localized within three different 6-DOF pose error thresholds: $ (0.25m, 2^{\circ})$, $(0.5m, 5^{\circ})$ and $(5m, 10^{\circ}) $ for high, medium and coarse precision respectively. We chose a structure-based localization method CSL\cite{svarm2016city} as well as two image-based localization methods FAB-MAP\cite{cummins2008fab} and NetVLAD\cite{arandjelovic2016netvlad} as baselines. NetVLAD is the best learning-based method for image retrieval and only second to DenseVLAD \cite{torii201524} which is currently the best image-based localization technique, but uses hand-crafted features. CSL attains higher localization accuracy than NetVLAD for the high- and medium-precision regimes, especially in the urban parts of the dataset.

\subsection{Validation of DIFL and FCL}
\label{sec4sub2}

In order to validate domain-invariant feature learning, we train the original multi-domain image translation model without feature consistency loss on the urban part of dataset with 12 domains and hyperparameter $ \lambda_{1}=10 $. We observe that the feature distance (\hyperref[feature_loss]{\ref{feature_loss}}) stabilizes at epoch 300 when using a learning rate of $ \alpha=0.0002 $, and so we linearly decrease the learning rate to zero during the next 300 epochs.

To train the model with feature consistency loss, we use transfer learning to fine tune the original model at epoch 300 after adding in the FCL. This is due to our experimental observations that it is difficult to successfully train the image translation model if we add in FCL starting from the first epoch. We also increase the hyperparameter $ \lambda_{2} $ from $ 0.05 $ to $ 0.1 $ linearly during the next 300 epochs of training. The configuration is the default we use for DIFL with FCL unless stated otherwise.
\begin{table}[h]
\caption{Ablation study for DIFL with FCL}
\label{tab1}
\begin{center}
\begin{tabular}{cccc|ccc}
		\hline
		\multirow{2}{*}{Train}&\multirow{2}{*}{$\lambda_{2}$}&\multirow{2}{*}{Test}&\multirow{2}{*}{PCA}& \multicolumn{3}{c}{Localization Accuracy (\%)}   \\ 
		& & & &    0.25m  2$ ^{ \circ } $  & 0.5m  5$ ^{ \circ } $ & 5m  10$ ^{ \circ } $  \\
		\hline
		\hline
	       ---& 0.0 & $ cosine $ &--- &\bf20.3&44.6&85.8 \\
	       ---& 0.0 & $ L2 $ &--- &19.5&43.4&83.7 \\
	    \hline
	       $ cosine $& 1.0 & $ L2 $ &--- &19.4&42.8&82.7 \\
	       $ cosine $& 1.0 & $ cosine $ &--- &20.0&44.4&86.4 \\
	       $ L2 $& 0.1 & $ L2 $ &--- &19.3&42.8&82.5 \\
	       $ L2 $& 0.1 & $ cosine $ &--- &20.2&\bf45.0&\bf87.2 \\
	       $ L2 $& 1.0 & $ L2 $ &--- &13.5&29.4&65.1 \\
	       $ L2 $& 1.0 & $ cosine $ &--- &15.5&34.8&77.6 \\
	    \hline
	       $ L2 $& 0.1 & $ L2 $ &slice &19.3&42.8&82.5 \\
	       $ L2 $& 0.1 & $ cosine $ &slice &20.2&\bf45.0&86.6 \\
	       $ L2 $& 0.1 & $ L2 $ &100 &14.0&31.3&78.5 \\
	       $ L2 $& 0.1 & $ cosine $ &100 &17.2&38.&83.7 \\
		
		\hline
	\end{tabular}
\end{center}
\end{table}
\begin{table}[h]
	\caption{Results in Comparison to Baselines}
	\label{tab2}
	\begin{center}
		\begin{tabular}{c|ccc|ccc|ccc}
			\hline
			\multirow{3}{*}{Method}& \multicolumn{3}{c|}{Urban(\%)} &\multicolumn{3}{c|}{Suburban(\%)} &\multicolumn{3}{c}{Park(\%)} \\
		&  \multicolumn{3}{c|}{.25/.50/5.0} &  \multicolumn{3}{c|}{.25/.50/5.0}&  \multicolumn{3}{c}{.25/.50/5.0}  \\ & \multicolumn{3}{c|}{2/5/10} & \multicolumn{3}{c|}{2/5/10}& \multicolumn{3}{c}{2/5/10} \\
			\hline
			CSL\cite{svarm2016city}& \multicolumn{3}{c|}{\textbf{36.7}/42.0/53.1} &  \multicolumn{3}{c|}{8.6/11.7/21.1}&  \multicolumn{3}{c}{7.0 /9.6 /17.0} \\
			FAB-MAP\cite{cummins2008fab}& \multicolumn{3}{c|}{2.7 /6.4 /27.3} &  \multicolumn{3}{c|}{0.5/1.5 /13.6}&  \multicolumn{3}{c}{0.8 /1.7 /11.5} \\
			NetVLAD\cite{arandjelovic2016netvlad}& \multicolumn{3}{c|}{17.4/40.3/\textbf{93.2}} &  \multicolumn{3}{c|}{7.7/21.0/\textbf{80.5}}&  \multicolumn{3}{c}{5.6 /15.7/65.8} \\
			\hline
			\bf DIFL(ours)& \multicolumn{3}{c|}{20.3/44.6/85.8} &  \multicolumn{3}{c|}{\textbf{9.2}/23.2/66.9}&  \multicolumn{3}{c}{\textbf{10.3}/26.3/69.6} \\
		\bf DIFL+FCL(ours)& \multicolumn{3}{c|}{20.2/\textbf{45.0}/87.2} &  \multicolumn{3}{c|}{9.1/\textbf{23.3}/69.4}&  \multicolumn{3}{c}{10.1/\textbf{26.4}/\textbf{74.0}} \\
			\hline 															
		\end{tabular}
	\end{center}
\end{table}

Additionally, considering DIFL without FCL is not efficient and may not fully converge at 600 epochs, we train it again with a constant learning rate during the first 600 epochs and then linearly decrease it from 0.0002 to 0 in the next 600 epochs. The results of localization using the three configurations above are shown in Figure \hyperref[fig4]{\ref{fig4}}. Note that they share the same training process in the first 300 epochs. The configuration is the default we use for DIFL without FCL unless stated otherwise.

We use $ cosine\  distance $ as the metric for testing and $ L2\ distance $ as FCL for training in the above experiments. Figure \hyperref[fig4]{\ref{fig4}} shows the efficiency and effectiveness of FCL for DIFL, where it is able to achieve higher accuracy with less training epochs, which is consistent with the claim put forth in Section \hyperref[sec3sub2]{\ref{sec3sub2}}. 

For the ablation study, Table \hyperref[tab1]{\ref{tab1}} shows how the results are influenced by using different distance metrics for training and testing, values of hyperparameters for FCL and dimensionality of PCA applied to domain-invariant feature before retrieval. The "Train" column shows the distance metric for FCL training, and is null if it is trained without FCL for the first two rows. The "Test" column shows the distance metric for testing during the image retrieval process. The "PCA" column shows the dimension the of feature vector after PCA. A value of "slice" in the "PCA" column indicates that the dimension is the number of images per slice.

From Table \hyperref[tab1]{\ref{tab1}}, we can see that testing with $ cosine\ distance $ is more effective for image retrieval regardless of the training process, and training without FCL benefits the high-precision localization but sacrifices accuracy on medium- and coarse-precision regimes. Applying PCA does not improve the result and achieves lower accuracy on the course-precision regime due to the dimensionality reduction.

\subsection{Results Comparison with Baselines}
\label{sec4sub3}

Table \hyperref[tab2]{\ref{tab2}} shows the comparison with several baselines, where the results of our baselines come form \cite{sattler2018benchmarking}. Our proposed methods achieve higher accuracy than baselines on the park part of the dataset in every precision regime. On the suburban and urban part, DIFL and FCL outperform NetVLAD in the high- and medium-precision regime. Additionally, our results are even better than the structure-based method CSL on the suburban and park part, where NetVLAD fails in the high-precision regime.

Overall, our proposed method does better in images from suburban and park parts, where foliage and vegetation appear more commonly and the domain-invariant feature is robust to the foliage variance across domains. Furthermore, the vehicles and pedestrians in images from the urban part hinder correct feature abstraction in DIFL+FCL, resulting in weaker localization results. The differing performance in different precision regimes could be due to the fact that DIFL was trained using images with the same pose but under different environments, leading to our network being more robust to environmental changes but less to perspective shifts. In the end, this causes better results for high- and medium-precision localization but worse results for course-precision compared to NetVLAD.

\section{CONCLUSIONS}
\label{sec5}

In this work, we have proposed a novel domain-invariant feature learning approach based on the ComboGAN architecture for a retrieval-based localization task. Our method is supposed to be robust to environmental condition changes. We formulate our model, propose feature consistency loss, and validate our approach on the challenging CMU-Seasons dataset, with comparison with several localization methods. Our results outperform the best learning-based methods for image retrieval in high- and medium-precision regimes, especially in park-line or otherwise high-foliage environments.

Our approach has presented promising results, especially with regards to generating domain-invariant features in latent space, which could be utilized in future works to estimate camera pose for long-term SLAM. However, one concern with our approach for image retrieval is that it is not very robust to dynamic objects in urban areas or huge perspective changes, which is an area that we hope to improve on in the future. 

\addtolength{\textheight}{-12cm}   








\bibliographystyle{IEEEtran}
\bibliography{IEEEabrv,mylib}

\end{document}